\documentclass[10pt,twocolumn,letterpaper]{article}

\usepackage[pagenumbers]{cvpr} 

\usepackage{booktabs}
\usepackage{multirow}
\usepackage{makecell}
\usepackage[table]{xcolor}
\usepackage{graphicx}
\usepackage{amsmath}
\usepackage{subcaption}

\definecolor{mygray}{gray}{.92}
\definecolor{mygreen1}{RGB}{253, 244, 244}
\definecolor{mygreen2}{RGB}{247, 247, 252}
\definecolor{ForestGreen}{RGB}{34,139,34}
\definecolor{Forestred}{RGB}{220,50,50}
\definecolor{seen_back}{RGB}{224, 241, 239}
\definecolor{unseen_back}{RGB}{243, 231, 213}

\definecolor{cvprblue}{rgb}{0.21,0.49,0.74}
\usepackage[pagebackref,breaklinks,colorlinks,allcolors=cvprblue]{hyperref}

\newcommand{\gold}[1]{\textcolor{red}{\textbf{#1}}}
\newcommand{\silver}[1]{\textcolor{blue}{\textbf{#1}}}

\newcommand{\tr}[1]{\scriptsize{\textcolor{darkgray}{\textit{#1}}}}
\def\paperID{3314} 
\def\confName{CVPR}
\def\confYear{2026}

\title{
Vision-Language Attribute Disentanglement and Reinforcement \\ for Lifelong Person Re-Identification
}

\author{Kunlun Xu\textsuperscript{\rm 1}\thanks{Equal contribution}~~~ Haotong Cheng\textsuperscript{\rm 1}\textsuperscript{\rm *}~~~ Jiangmeng Li\textsuperscript{\rm 2}~~~ Xu Zou\textsuperscript{\rm 3}~~~ Jiahuan Zhou\textsuperscript{\rm 1}\thanks{Corresponding author}  \\
 {\small \textsuperscript{\rm 1} Wangxuan Institute of Computer Technology, Peking University, Beijing, China}\\
 {\small \textsuperscript{\rm 2} University of Chinese Academy of Sciences, Beijing, China}\\
 {\small \textsuperscript{\rm 3} School of Artificial Intelligence and Automation, Huazhong University of Science and Technology, Wuhan, China}\\
 {\small xkl@stu.pku.edu.cn~~~~chenght9923@mails.jlu.edu.cn~~~~jiangmeng2019@iscas.ac.cn~~~~zoux@hust.edu.cn~~~~jiahuanzhou@pku.edu.cn}\\
}

\begin{document}
\maketitle
\begin{abstract}
Lifelong person re-identification (LReID) aims to learn from varying domains to obtain a unified person retrieval model. Existing LReID approaches typically focus on learning from scratch or a visual classification-pretrained model, while the Vision-Language Model (VLM) has shown generalizable knowledge in a variety of tasks. Although existing methods can be directly adapted to the VLM, since they only consider global-aware learning, the fine-grained attribute knowledge is underleveraged, leading to limited acquisition and anti-forgetting capacity. To address this problem, we introduce a novel VLM-driven LReID approach named Vision-Language Attribute Disentanglement and Reinforcement (VLADR). Our key idea is to explicitly model the universally shared human attributes to improve inter-domain knowledge transfer, thereby effectively utilizing historical knowledge to reinforce new knowledge learning and alleviate forgetting. Specifically, VLADR includes a Multi-grain Text Attribute Disentanglement mechanism that mines the global and diverse local text attributes of an image. Then, an Inter-domain Cross-modal Attribute Reinforcement scheme is developed, which introduces cross-modal attribute alignment to guide visual attribute extraction and adopts inter-domain attribute alignment to achieve fine-grained knowledge transfer. Experimental results demonstrate that our VLADR outperforms the state-of-the-art methods by 1.9\%-2.2\% and 2.1\%-2.5\% on anti-forgetting and generalization capacity.
Our source code is available at 
\href{https://github.com/zhoujiahuan1991/CVPR2026-VLADR}{https://github.com/zhoujiahuan1991/CVPR2026-VLADR}.
\end{abstract}

\begin{figure}[t]
    \centering
    \vspace{-10pt}
    \includegraphics[width=1\linewidth]{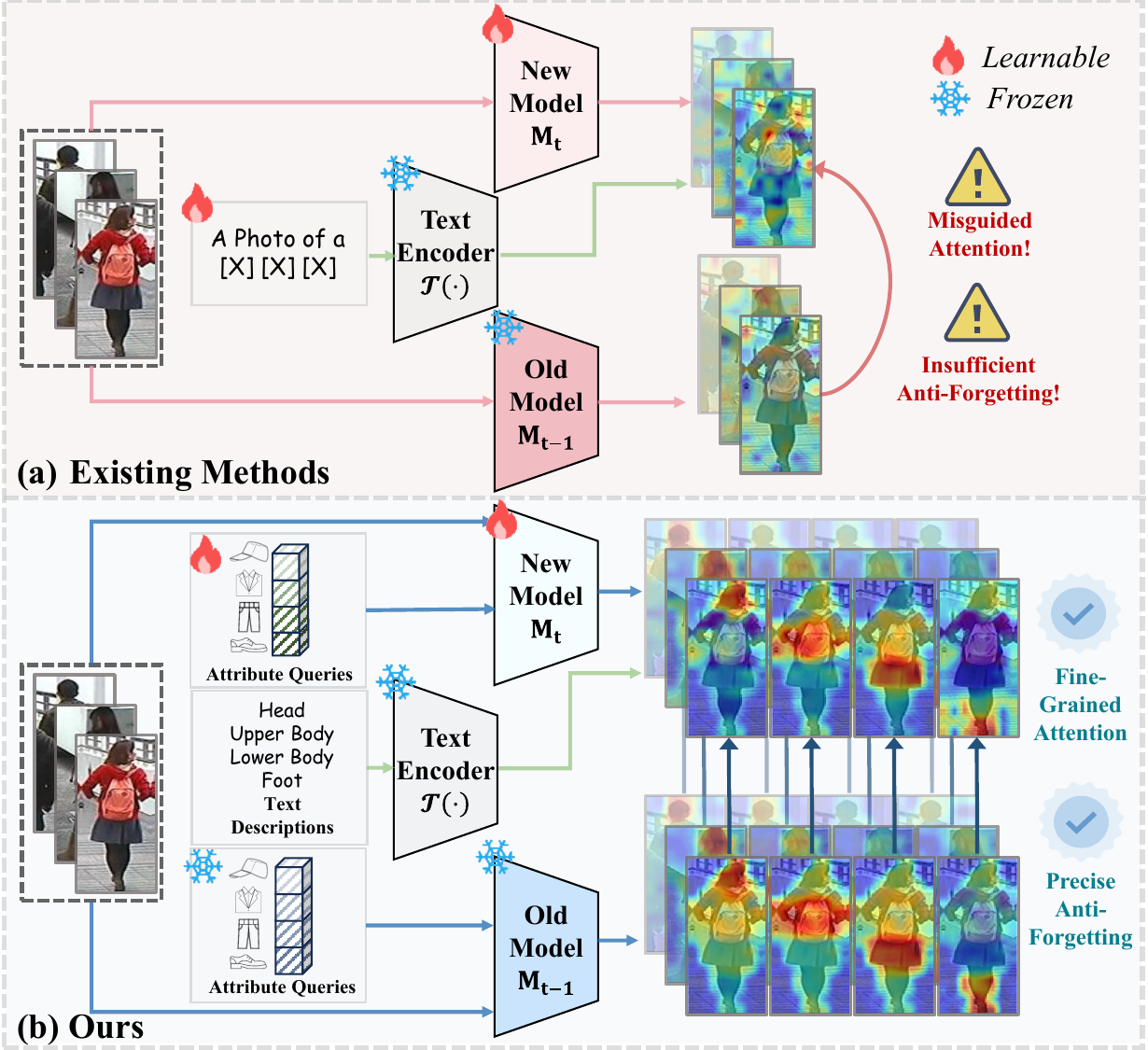}
    \vspace{-10pt}
    \caption{(a) Existing methods rely on global feature learning when exploiting VLM, suffering from misguided attention on background and insufficient utilization of the human semantics. (b) Our method explicitly models and transfers human attribution knowledge across domains, achieving more precise human semantic acquisition and continual attribute knowledge reinforcement. }
     \vspace{-10pt}
    \label{fig:first}
\end{figure}

\vspace{-10pt}
\section{Introduction}
\label{sec:intro}
Person re-identification (ReID) aims to retrieve the same individual across non-overlapping camera views~\cite{gong2022person,cui2024dma,shi2023dual}. While conventional ReID methods primarily address the stationary scenarios~\cite{gong2024cross},  real-world deployments often operate in dynamic settings where data continuously stream in under varying weather, illumination, and resolution domains~\cite{wu2021generalising,xu2025componential,zhang2025scap,ge2022lifelong,xu2025long}. To close the gap, lifelong person re-identification (LReID) ~\cite{xu2024distribution,xu2024mitigate,cui2024learning} has emerged as a promising research direction. The primary challenge in LReID lies in the continuous integration of domain-specific knowledge to construct a universal model that remains robust and adaptable to evolving conditions~\cite{xu2024lstkc,cui2025dkc}.

Existing LReID approaches primarily follow a conventional learning paradigm, in which models are either trained from scratch or initialized with a visually pre-trained classification backbone~\cite{pu2021lifelong,li2024exemplar}. However, owing to the lack of human-centric prior knowledge, these methods often suffer from limited robustness and generalization capability~\cite{hu2024empowering}.

Recent advances have demonstrated remarkable generalization capabilities of Vision-Language Models (VLMs) across various visual tasks, including ReID~\cite{li2023clip,hu2024empowering,yang2024pedestrian}. Although existing LReID methods can be integrated into a VLM-based baseline, as shown in Fig.~\ref{fig:first} (a), they predominantly rely on global-aware learning with coarse supervision. Such supervision often causes the model to overfit to irrelevant regions, such as background objects, instead of focusing on person-specific regions. Consequently, these learning paradigms misguide the knowledge accumulation process and hinder the acquisition of new information, ultimately degrading model robustness.

To address these limitations, we propose a novel framework termed \textbf{V}ision–\textbf{L}anguage \textbf{A}ttribute \textbf{D}isentanglement and \textbf{R}einforcement (VLADR). As illustrated in Fig.~\ref{fig:first} (b), the central idea of VLADR is to disentangle universally shared human attributes to achieve more precise and human-centric knowledge representation. This design enables effective transfer of human-relevant historical knowledge to new domains, thereby continually reinforcing fine-grained attribute knowledge. Specifically, VLADR consists of two main components: a Multi-grain Text Attribute Disentanglement (MTAD) mechanism and an Inter-domain Cross-modal Attribute Reinforcement (ICAR) scheme. MTAD extracts both global and diverse local textual attributes from images based on a pretrained VLM. Building upon these disentangled attributes, ICAR conducts cross-modal attribute alignment to enable fine-grained visual attribute learning. In addition, an inter-domain attribute alignment strategy is introduced to facilitate attribute knowledge transfer and retention across domains. Through the joint operation of MTAD and ICAR, human-relevant attribute knowledge is progressively reinforced throughout sequential training. Extensive experiments demonstrate that VLADR substantially enhances both knowledge acquisition and knowledge consolidation in LReID.

    In summary, the contributions of this work are threefold: 
(1) We present a pioneering study that leverages VLMs for lifelong person re-identification and propose the novel VLADR framework that effectively utilizes multi-granularity human-centric knowledge to facilitate knowledge acquisition while mitigating catastrophic forgetting. (2) We develop a Multi-grain Text Attribute Disentanglement (MTAD) mechanism that extracts diverse coarse- and fine-grained textual attributes, coupled with an Inter-domain Cross-modal Attribute Reinforcement (ICAR) scheme that continuously reinforces visual attribute knowledge across domains throughout the lifelong learning process. (3) Extensive experiments demonstrate that our VLADR achieves state-of-the-art performance.

\section{Related Work}
\subsection{Lifelong Person Re-Identification}

Person Re-Identification (ReID) aims to match individuals across non-overlapping camera views. While conventional methods predominantly operated under the assumption that training and testing data originated from the same domain~\cite{li2018HAN, luo2019bag}. This premise significantly limited their practicality in real-world surveillance systems, where environmental conditions are inherently dynamic~\cite{zhuang2020rethink, he2021transreid, chen2018camera}. To overcome this limitation, Lifelong Person Re-Identification (LReID) has emerged as a promising paradigm, aiming to learn continuously from a sequential stream of domains. The central challenge in LReID is catastrophic forgetting, a tendency where models overwrite old knowledge when learning from new domains~\cite{li2018lwf, Shemelkov2017IL}. Existing LReID approaches could be broadly categorized into exemplar-based and exemplar-free methods~\cite{xu2024lstkc}.

Exemplar-based methods alleviated forgetting by storing representative samples from past domains (exemplars) and replaying them during subsequent training~\cite{wu2021generalising, ge2022lifelong, yu2023lifelong}. Although these methods often yielded strong performance, they incurred linearly increasing storage and computational overhead. More critically, the storage of raw human images raised substantial privacy concerns, hindering their deployment in real-world scenarios~\cite{zhou2025distribution, xu2024mitigate}. Consequently, our work focuses on the more challenging and practical exemplar-free setting.

Existing exemplar-free methods primarily mitigated forgetting through techniques such as knowledge distillation, prototype learning, and style transfer. Knowledge distillation-based approaches enforced consistency between historical and current model outputs ~\cite{pu2021lifelong,xu2024lstkc,xu2025long,sun2022patch}. Prototype-based methods~\cite{xu2024distribution,zhou2025distribution} introduced prototypes to preserve historical information. Style transfer-based techniques replicated visual styles from previous domains to enhance knowledge retention~\cite{xu2025dask}. However, these methods typically either learned from scratch or initialized with a generic classification-pretrained model~\cite{pu2021lifelong,cui2023dcr}, thereby lacking explicit human-centric prior knowledge. This fundamental limitation restricted their capacity to learn discriminative representations robust to domain shifts and long-term forgetting~\cite{wang2022dualprompt}.

\begin{figure*}[htbp]
    \centering
    \includegraphics[width=1.0\linewidth]{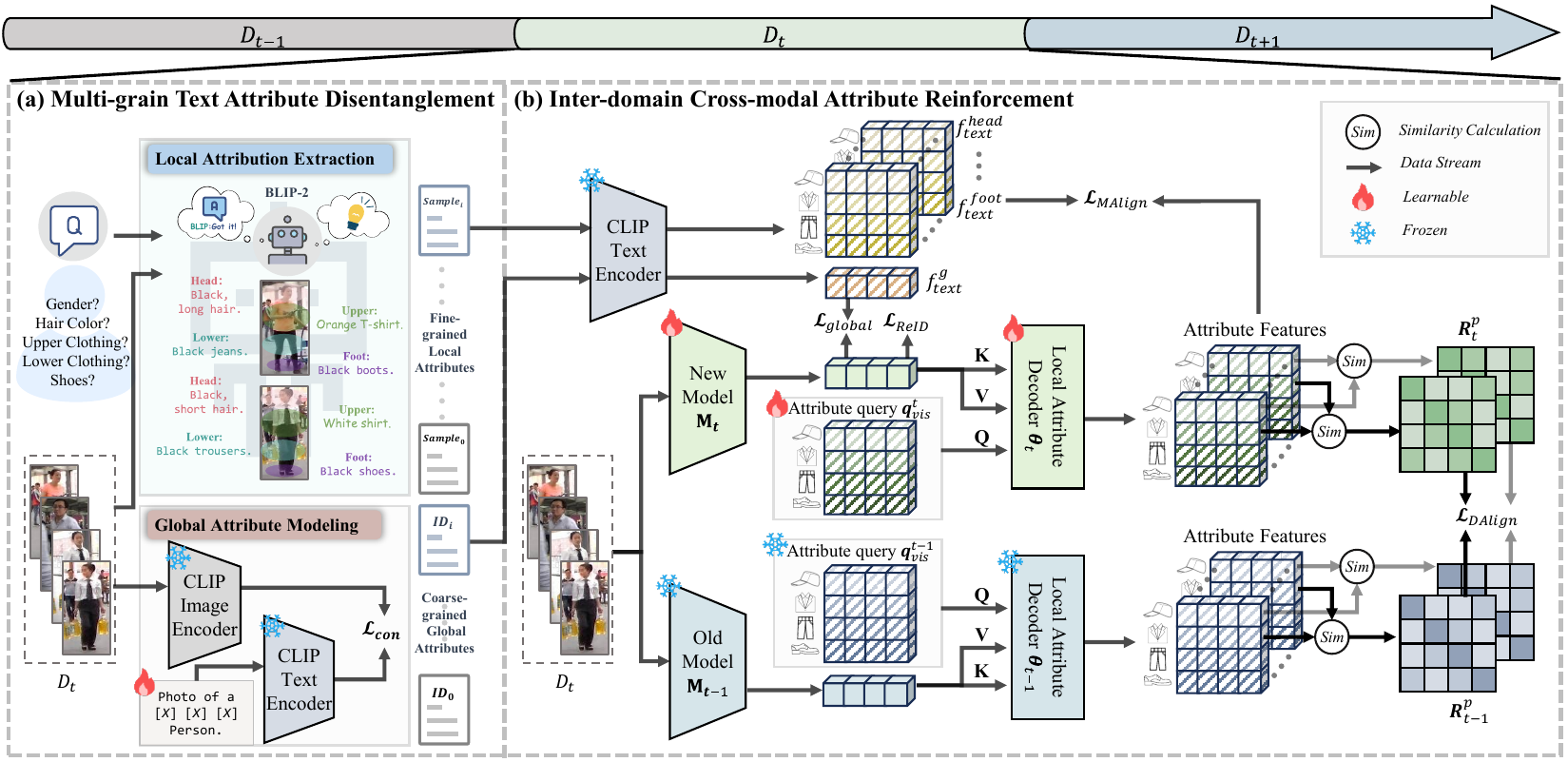}
    \caption{Overview of our framework. Given the new data $D_t$, two stages operate sequentially. (a) Multi-grain Text Attribute Disentanglement mechanism conducts local attribution extraction and a global attribute model based on pretrained VLMs. Then, these disentangled multi-grain attributes are utilized by (b) Inter-domain Cross-modal Attribute Reinforcement scheme, which conducts visual-text alignment $\mathcal{L}_{MAlign}$ to guide visual attribute extraction. Besides, the inter-domain attribute alignment $\mathcal{L}_{DAlign}$ is also performed to accumulate the continually learned knowledge. }
    \label{fig:framework}
\end{figure*}

\subsection{VLM-based Person Re-Identification}
VLMs acquired extensive real-world knowledge during pre-training and have demonstrated significant potential in enhancing various vision tasks~\cite{zhang2025scap,mao2023clip4hoi}. In light of this, several works have attempted to leverage VLM for person re-identification~\cite{li2023clip,yang2024pedestrian,hu2024empowering,chen2023unveiling}. These methods typically imposed a global alignment constraint during downstream ReID training, maintaining the original vision-language alignment architecture. However, such coarse-grained constraints failed to guide the model to precisely and comprehensively extract the human-relevant information. As a consequence, the noise-prone features accumulated progressively during lifelong learning, which disrupted effective knowledge acquisition and undermined model robustness~\cite{xu2024lstkc,xu2024mitigate}. Moreover, incompletely learned features limited the utilization of new training data. To overcome these limitations, we investigate guiding the model in learning diverse human attributes grounded in VLM knowledge, thereby facilitating both robust feature representation and reliable long-term knowledge accumulation.

\section{The Proposed Method}
\subsection{Problem Definition}
LReID addresses the scenarios where a stream of training datasets $\mathcal{D}=\{D_t\}_{t=1}^{T}$ are provided incrementally, and each dataset $D_t=\{(x_i,y_i)\}_{i=1}^{n_t}$ consists of $n_t$ image-label pairs with $x_i$ being an image and $y_i$ its corresponding identity label. When $D_t$ is presented for training, all previous datasets $D_1,...,D_{t-1}$ are inaccessible~\cite{xu2024distribution}. After the $t$-th training step, the final  model is denoted as $\boldsymbol{\mathrm{M}}_t$, which is evaluated on $T$ test datasets $\mathcal{D}^{te}=\{D_t^{te}\}_{t=1}^{T}$  drawn form  $T$ identical training domains. Additionally, $U$ unseen datasets $\mathcal{D}^{un}=\{ D^{un}_t\}_{t=1}^{U}$ collected from novel domains are employed to evaluate the generalization capacity of $\boldsymbol{\mathrm{M}}_T$.

\subsection{Overview}
As illustrated in Fig.~\ref{fig:framework}, our framework consists of two core stages: Multi-grain Text Attribute Disentanglement (MTAD) and Inter-domain Cross-modal Attribute Reinforcement (ICAR). Given incoming data $D_t$, MTAD first leverages BLIP to generate part-wise textual attribute descriptions for each instance, while simultaneously utilizing  CLIP to extract ReID-aware global attributes. Subsequently, ICAR encodes these multi-grain textual attribute features through the CLIP text encoder. Visual global and local attribute features are then extracted and aligned with their textual counterparts through cross-modal alignment. Additionally, to facilitate knowledge accumulation across domains, an inter-doamin attribute knowledge alignment mechanism is developed to achieve knowledge trnasfer.

\subsection{Multi-grain Text Attribute Disentanglement}\label{sec:MTAD}
In standard LReID benchmarks, only the human images and the corresponding identity labels are provided. Therefore, to effectively leverage the semantic knowledge in VLMs, generating an appropriate textual description becomes crucial. 
In this section, we introduce how to generate the local and global text attributes without using auxiliary data.

\textbf{Local Attribution Extraction}. Recent advancements in image-to-text generation models have demonstrated remarkable capabilities for producing human descriptions. However, existing approaches typically generate only a single paragraph or phrase per sample~\cite{yang2024mllmreid,wang2024large}. Such strategies fail to capture fine-grained information, leading to limited knowledge acquisition capacity, as shown in Fig.~\ref{fig:first} (a). 

To address this issue, we disentangle the human body into distinct local semantic regions, including head, upper body, lower body and foot. A set of textual queries $\mathcal{Q}=\{T^Q_i\}_{i=1}^{N_a}$ are predefined, with each targeting a specific local attribute. Given an input image, the corresponding attribute $T_i^A$ is extracted by the BLIP model $\mathcal{B}$ as following:
\begin{equation}
    T_{x,i}^A=\mathcal{B}(x,T_i^Q)
    \label{eq:blip}.
\end{equation}

Consequently, a set of local textual semantic attributes $\mathcal{T}_x=\{T^A_{x,i}\}_{i=1}^{N_a}$ is obtained according to Eq.~\ref{eq:blip}. $\mathcal{T}_x$ provides a precise description of human parts. It is worth noting that excessively fine-grained local attributes do not necessarily yield better performance, and the detailed configurations of selected local attributes and $\mathcal{Q}$ are provided in our Supplementary Materials.

\textbf{Global Attribute Modeling}. In addition to local attributes, we introduce global attribute modeling to provide complementary semantic information, enabling more comprehensive human characterization. Inspired by~\cite{li2023clip}, we extract the ReID-aware global attribute based on the CLIP model. Specifically, a set of learnable text prompts $\{[X]_i\}_{i=1}^M$ are assigned to each identity. The input image $x$ and the corresponding text description ``A photo of a $[X]_1[X]_2[X]_3\dots[X]_M$'' are fed into the frozen image encoder $\mathcal{I(\cdot)}$ and text encoder $\mathcal{T}(\cdot)$, respectively. The text prompts are optimized by minimizing:
\begin{equation}
    \mathcal{L}_{prompt}=\sum_x\mathcal{L}_{i2t}(x)+\sum_x\mathcal{L}_{t2i}(x),
    \label{eq:stage1_loss}
\end{equation}
where $\mathcal{L}_{i2t}$ and $\mathcal{L}_{t2i}$ denote the contrastive loss items as depicted in~\cite{li2023clip}. 
To preserve the generalizable knowledge in CLIP, both $\mathcal{I(\cdot)}$ and $\mathcal{T(\cdot)}$ remain frozen as the training stage $t$ increases.

\subsection{Inter-domain Cross-modal Attribute Reinforcement} 
In this section, we aim to mine and accumulate visual knowledge throughout the lifelong learning process. Specifically, three learning branches are introduced: Coarse-grained Global Representation Learning, Fine-grained Local Attribute Mining, and Inter-domain Attribute Knowledge Transfer.

\textbf{Coarse-grained Global Representation Learning}. 
Given a batch of $B$ input samples $\{(x_i, y_i)\}_{i=1}^B$, we employ a ViT-based visual model $\mathbf{M}_t$ to extract features from each instance $x_i$. Specifically, $x_i$ is tokenized~\cite{li2024exemplar} into a sequence represented as $\boldsymbol{f}^i_t\in\mathbb{R}^{L_h\times d}$, where $L_h$ and $d$ are the sequence length and token dimension, respectively. Additionally, a [CLS] token $\boldsymbol{f}_{cls}$ is attached to $\boldsymbol{f}_t^i$, resulting in a sequence $\boldsymbol{h}_t^i=[\boldsymbol{f}_t^i,\boldsymbol{f}_{cls}]$. After processed by $\mathbf{M}_t$, $\boldsymbol{f}_{cls}$ is treated as the global vison feature $\boldsymbol{f}^{g,i}_{vis}$. Following previous works~\cite{xu2024distribution,xu2024lstkc}, the classical ReID loss $\mathcal{L}_{ReID}$ is adopted to ensure knowledge learning, which is defined as:
\begin{equation}
\mathcal{L}_{ReID}=\mathcal{L}_{Tri}+\mathcal{L}_{ce},
\end{equation}
where $\mathcal{L}_{Tri}$ and $\mathcal{L}_{ce}$ are the triplet loss  ~\cite{huang2022lifelong} and a cross-entropy loss~\cite{xu2024distribution}.

Subsequently, the text prompt corresponding to identity $y_i$ is utilized to generate the textual global attribute feature $\boldsymbol{f}^{g,i}_{text}$. To establish cross-modal correspondence, the visual-text global attribute alignment loss is conducted:
\begin{equation}
    \mathcal{L}_{global}=-\frac{1}{B}\sum_{i=1}^B\log\frac{\exp(s(\boldsymbol{f}^{g,i}_{vis}, \boldsymbol{f}^{g,i}_{text}))}{\sum_{k=1}^K\exp(s(\boldsymbol{f}^{g,i}_{vis}, \boldsymbol{f}^{g,k}_{text}))},
    \label{eq:coarse}
\end{equation}
where $s(\cdot,\cdot)$ is the cosine similarity.

\textbf{Fine-grained Local Attribute Mining.} While MTAD enables the extraction of multi-grain textual attributes, corresponding visual attributes still remain unavailable due to the prohibitive annotation cost in LReID settings. To overcome this limitation, we introduce an adaptive visual attribute learning paradigm. Specifically, a set of $N_a$ learnable attribution query $\boldsymbol{q}_{vis}\in\mathbb{R}^{N_a\times d}$ is defined, each corresponding to a local attribute defined in Sec.~\ref{sec:MTAD}. 

Then, a Local Attribute Decoder $\boldsymbol{\theta}_t$, implemented as a cross attention layer, is employed to decode the visual attributes. Specifically, the image token sequence $\boldsymbol{f}^i_t$ serves as the Key ($\boldsymbol{K}$) and value ($\boldsymbol{V}$). $\boldsymbol{q}_{vis}$ is adopted as Query ($\boldsymbol{Q}$). The visual local attribute features $F_{vis}^{l,t}$ are thus computed as:
\begin{equation}
F_{vis}^{l,t}=\boldsymbol{\theta}_t(\boldsymbol{q}_{vis},\boldsymbol{f}^i_t,\boldsymbol{f}^i_t).
\end{equation}

On the textual side, the CLIP text encoder processes the local attributes $\mathcal{T}_{x_i}$ generated in Sec.~\ref{sec:MTAD}, obtaining the textual local attribute features $F^{l,t}_{text}\in\mathbb{R}^{N_a\times d}$:
\begin{equation}
F^{l,t}_{text}=\mathcal{T}(\mathcal{T}_{x_i}).
\end{equation}

The visual-text local attribute alignment is then optimized through:
\begin{equation}
    \mathcal{L}_{MAlign}=\frac{1}{N_a}\sum_{p=1}^{N_a}\mathcal{L}_{cons}^p,
\end{equation}
where $\mathcal{L}_{cons}^p$ denotes the contrastive loss for single attribute. Inspired by ~\cite{li2023clip,liu2025tp}, $\mathcal{L}_{cons}^p$ is defined as:
\begin{equation}
\begin{split}
       \mathcal{L}_{cons}^p=-\frac{1}{2B}\sum_{i=1}^B\Big(&\log\frac{\exp(s(\boldsymbol{v}_i^{p,t},\boldsymbol{t}_i^{p,t}))}{\sum_{j=1}^B\exp(s(\boldsymbol{v}_i^{p,t},\boldsymbol{t}_j^{p,t}))}\\
       +&\log\frac{\exp(s(\boldsymbol{v}_i^{p,t},\boldsymbol{t}_i^{p,t}))}{\sum_{k=1}^B\exp(s(\boldsymbol{v}_k^{p,t},\boldsymbol{t}_i^{p,t}))}\Big) ,
\end{split}
\end{equation}
where $\boldsymbol{v}^{p,t}\in\mathbb{R}^{B\times d}$ and $\boldsymbol{t}^{p,t}\in\mathbb{R}^{B\times d}$ represent visual and textual features of attribute $p$ across the batch,  extracted from $F^{l,t}_{vis}$ and $F^{l,t}_{text}$ respectively. 
\begin{table*}[htbp]
   \centering
   \caption{ Comparison results of Seen-domain anti-forgetting and UnSeen-domain generalization on Training Order-1.}
  \setlength{\tabcolsep}{1.0mm}{
\begin{tabular}{clccccccccccc>{\columncolor{seen_back}}c>{\columncolor{seen_back}}c>{\columncolor{unseen_back}}c>{\columncolor{unseen_back}}c}
    \toprule
     \multirow{2}{*}{Net.} & \multirow{2}{*}{Method} & \multirow{2}{*}{Pub.} & \multicolumn{2}{c}{Market} & \multicolumn{2}{c}{SYSU} & \multicolumn{2}{c}{LPW} & \multicolumn{2}{c}{MSMT17} & \multicolumn{2}{c}{CUHK03} & \multicolumn{2}{>{\columncolor{seen_back}}c}{\textbf{Seen-Avg}} & \multicolumn{2}{>{\columncolor{unseen_back}}c}{\textbf{Unseen-Avg}} \\
     \cmidrule{4-17}
      & & & mAP & R@1 & mAP & R@1& mAP & R@1& mAP & R@1& mAP & R@1& mAP & R@1& mAP & R@1 \\
      \midrule
     \multirow{6}{*}{\rotatebox{90}{ResNet}} 
     &LwF\cite{li2018lwf}&\scriptsize{\textcolor{darkgray}{\textit{TPAMI 2017}}}& 44.4 & 68.7 & 66.6 & 69.8 & 25.0 & 36.0 &6.8 & 19.6 & 52.9 & 53.6 & 39.1 & 49.5 & 50.9 & 43.8 \\
     & DKP\cite{xu2024distribution} & \tr{CVPR 2024} & 60.0&80.3&	84.1&	85.9	&46.0&	57.9&	17.7&	38.5&	41.0&	41.4&	49.8&	60.8&	57.5&	50.7 \\
      & LSTKC\cite{xu2024lstkc} & \tr{AAAI 2024} &57.0 & 	78.6&	82.9&	84.9&	47.2&	58.4&	18.4&	41.1&	42.3&	43.7&	49.6&	61.3&	57.8&	50.2 \\
      & DKP++\cite{zhou2025distribution} & \tr{TPAMI 2025} & 52.1	&76.5&	75.6&	77.7&	45.2&	58.1&	14.9&	36.1	&40.1&	42.2&	45.6&	58.1&	56.6&	48.5 \\
      & LSTKC++\cite{xu2025long} & \tr{TPAMI 2025} & 67.5&	83.9&	86.3&	88.2	&52.1	&61.9&	19.4&	40.3&	47.0&	47.8	&54.5	&64.4&	62.9	&55.6 \\
      & DASK\cite{xu2025dask} & \tr{AAAI 2025} & 49.3&	74.0&	74.9&	77.0&	45.7&	58.6&	17.6&	42.0&	34.5&	35.6&	44.4&	57.5&	59.6&	52.5 \\
      \midrule
      \multirow{2}{*}{\rotatebox{90}{ViT}} & PAEMA\cite{li2024exemplar} & \tr{IJCV 2024} &73.1&88.0
      &90.0&91.5&
    59.5  &68.6&30.4&54.4
      &48.4&48.4&60.3
      &70.2&69.8&63.2
      \\
      & DRE\cite{liu2025diverse} & \tr{TNNLS 2025}&	61.7&	80.8&	88.0&	89.8&	52.6&	64.7&	32.0&	57.3&	59.4&	61.3&	58.7&	70.8&	62.9&	58.1 \\
      \midrule

      \multirow{8}{*}{\rotatebox{90}{CLIP}} & LwF\cite{li2018lwf}	& \tr{TPAMI 2017} &51.4  &77.6 
      &73.6&76.5&36.3
      &46.9&13.7&35.0
      &27.2&26.0&40.4
      &52.4&55.0&47.1 
      \\
      & DKP\textsuperscript{$\dagger$}\cite{xu2024distribution} & \tr{CVPR 2024} & 81.1&91.4&	90.0	&91.0&	66.9&	74.4&	37.4&	63.6&	59.5&	62.0&	67.0	&76.5&	72.5&	64.9 \\
 & LSTKC\textsuperscript{$\dagger$}\cite{xu2024lstkc} & \tr{AAAI 2024} & 79.8	&91.4&	92.0&	92.7&	65.8&	75.1&	37.8&	63.5&	60.6&	62.4&	67.2&	77.0&	75.3&	68.5 \\
 & PAEMA\textsuperscript{$\dagger$}\cite{li2024exemplar} & \tr{IJCV 2024} & 81.3
 & 92.1 & 92.2 & 93.0 & 66.9 & 75.4 & 39.2 & 65.4 & 58.3 & 60.3 & 67.6 & 77.2 & \textcolor{blue}{\textbf{75.6}}& \textcolor{blue}{\textbf{69.8}} \\
 & DKP++\textsuperscript{$\dagger$}\cite{zhou2025distribution} & \tr{TPAMI 2025} & 79.4	&90.5	&87.8&	88.5&	66.1	&74.6&	37.7	&64.2&	57.9	&59.3	&65.8&	75.4&	73.0	&66.2 \\
 & LSTKC++\textsuperscript{$\dagger$}\cite{xu2025long} & \tr{TPAMI 2025} & 69.3	&85.5	&89.9	&90.9&	60.3&	71.4	&38.3&	64.0	&69.1	&70.4&	65.4&	76.4&	74.1	&67.7 \\
 & DASK\textsuperscript{$\dagger$}\cite{xu2025dask} & \tr{AAAI 2025} & 79.7 & 90.7
 &91.0&91.7&64.7
 &73.9&42.1&67.9
 &63.4&65.5&\textcolor{blue}{\textbf{68.2}}
 &\textcolor{blue}{\textbf{78.0}}&74.2&66.7
 \\
 \cmidrule{2-17}
 & \textbf{VLADR}& \tr{This Paper} & 80.1	&91.9&	92.0&92.5&67.8&76.2&	45.5&	70.5&	66.6&	68.4&	\textcolor{red}{\textbf{70.4}}&	\textcolor{red}{\textbf{79.9}}&	\textcolor{red}{\textbf{77.8}}&	\textcolor{red}{\textbf{72.0}} \\
 \bottomrule
    \end{tabular}}
\raggedright
 \footnotesize{ Net. represents the Backbone network. $\dag$ indicates integrating the LReID approach with CLIP-ReID~\cite{li2023clip} baseline.}  \\    
    \label{tab:order1}
     
\end{table*}

\textbf{Inter-domain Attribute Knowledge Transfer.} 
To facilitate the transfer of  previously learned attribute knowledge to the current model, we leverage the frozen visual model $\mathbf{M}_{t-1}$, attribute query $\boldsymbol{q}_{vis}^{t-1}$, and local attribute decoder $\boldsymbol{\theta}_{t-1}$ to process each input sample $x_i$, generating the historical visual local attribute features $F^{l,t-1}_{vis}\in\mathbb{R}^{N_a\times d}$. Let  $\boldsymbol{v}^{p,t-1}\in\mathbb{R}^{B\times d}$ denote the historical features of attribute $p$ in a batch, extracted from $F^{l,t-1}_{vis}$. 
Following~\cite{xu2025dask}, the cross-instance relation matrices $\boldsymbol{R}_{t-1}^p=\gamma(\boldsymbol{v}^{p,t-1}(\boldsymbol{v}^{p,t-1})^\top)$ and $\boldsymbol{R}_{t}^p=\gamma(\boldsymbol{v}^{p,t}(\boldsymbol{v}^{p,t})^\top)$ are calculated, where $\gamma(\cdot)$ is a row-wise normalization operator. Finally, the inter-domain attribute alignment loss is then defined as
\begin{equation}
    \mathcal{L}_{DAlign}=\frac{1}{N_a}\sum_{p=1}^{N_a}\mathcal{L}_{Rel}^p
    \label{eq:SKD},
\end{equation}
where $\mathcal{L}_{Rel}^p$ is a Kullback-Leibler (KL) divergence-based relation distillation loss calculated by:
\begin{equation}
    \mathcal{L}_{Rel}^p=\frac{1}{B}\sum_{i=1}^BKL\Big((\boldsymbol{R}_t^p)_i ||(\boldsymbol{R}_{t-1}^p)_i\Big)
    \label{eq:SKD},
\end{equation}

\textbf{Training and Inference.} During the MTAD stage (Sec.~\ref{sec:MTAD}), only Eq. \eqref{eq:stage1_loss} is optimized. In the ICAR stage, the overall loss $\mathcal{L}$ is combined as:
\begin{equation}
\mathcal{L}=\mathcal{L}_{ReID}+\mathcal{L}_{global}+\alpha \mathcal{L}_{MAlign}+\beta \mathcal{L}_{DAlign},
\end{equation}
where $\alpha$ and $\beta$ are hyperparameters to balance different loss components. 

During inference, the final visual model $\mathbf{M}_T$, local attribute query $\boldsymbol{q}_{vis}^T$ and local attribute decoder $\boldsymbol{\theta}_T$ are adopted for feature extraction. Specifically, given an input image $x$, the global feature $f_{vis}^g$ extracted by $\mathbf{M}_T$ and the visual local attribute features $F_{vis}^{l,T}$ extracted by $\boldsymbol{\theta}_T$ and $\boldsymbol{q}_{vis}^T$, form the final feature:
\begin{equation}
   f_{vis}^c=[f_{vis}^g,F_{vis}^{l,T}].
\end{equation}
This comprehensive feature vector  $f_{vis}^c$ is subsequently adopted for person retrieval.

\begin{table*}[t]
   \centering
   \caption{ Comparison results of Seen-domain anti-forgetting and UnSeen-domain generalization on Training Order-2.}
  \setlength{\tabcolsep}{1.0mm}{
\begin{tabular}{clccccccccccc>{\columncolor{seen_back}}c>{\columncolor{seen_back}}c>{\columncolor{unseen_back}}c>{\columncolor{unseen_back}}c}
    \toprule
     \multirow{2}{*}{Net.} & \multirow{2}{*}{Method} & \multirow{2}{*}{Pub.} & \multicolumn{2}{c}{LPW} & \multicolumn{2}{c}{MSMT17} & \multicolumn{2}{c}{Market} & \multicolumn{2}{c}{SYSU} & \multicolumn{2}{c}{CUHK03} & \multicolumn{2}{>{\columncolor{seen_back}}c}{\textbf{Seen-Avg}} & \multicolumn{2}{>{\columncolor{unseen_back}}c}{\textbf{Unseen-Avg}} \\
     \cmidrule{4-17}
      & & & mAP & R@1 & mAP & R@1& mAP & R@1& mAP & R@1& mAP & R@1& mAP & R@1& mAP & R@1 \\
      \midrule
     \multirow{6}{*}{\rotatebox{90}{ResNet}} 
     &LwF\cite{li2018lwf}& \tr{TPAMI 2017} & 38.8 & 50.5 & 6.7 & 19.2 & 43.9 & 67.1 & 73.2 & 75.9 & 50.1 & 51.9 & 42.5 & 52.9 & 50.0 & 42.7\\
     & DKP\cite{xu2024distribution} & \tr{CVPR 2024} &49.5 &61.4 &14.1 &32.6 &60.3 &80.6 &84.5 &86.4 &43.6 &43.7 &50.4 &60.9 &59.5 &52.4 \\
      & LSTKC\cite{xu2024lstkc} & \tr{AAAI 2024} & 46.7& 57.6& 14.9& 33.9& 56.5& 78.0& 84.0& 86.1& 42.1& 43.7& 48.8& 59.9& 57.4& 49.5\\
      & DKP++\cite{zhou2025distribution} & \tr{TPAMI 2025} &52.2 &64.0 &18.0 &41.0 &63.7 &82.5 &81.3 &82.9 &49.7 &51.8 &53.0 &64.4 &61.0 &54.0 \\
      & LSTKC++\cite{xu2025long} & \tr{TPAMI 2025} & 55.7&65.7 &15.0 &34.3 &66.9 &83.6 &86.3 &87.6 &48.4 &49.9 &54.4 &64.2 &63.4 &56.0 \\
      & DASK\cite{xu2025dask} & \tr{AAAI 2025} & 47.2& 60.4& 16.1& 39.9& 58.7& 79.9& 77.5&79.4 &42.0 &43.0 &48.3 &60.5 &60.0 &52.8 \\
      \midrule
      \multirow{2}{*}{\rotatebox{90}{ViT}} & PAEMA\cite{li2024exemplar} & \tr{IJCV 2024} &64.7 &73.9 &26.5 &50.0 &73.5 &87.6 &90.6 &91.5 &48.9 &49.4 &60.9 &70.5 &70.5 &63.8 \\
      & DRE\cite{liu2025diverse} & \tr{TNNLS 2025} & 42.2&52.3 &23.6 &45.8 &68.2 &84.3 &87.3 &89.1 &51.9 &53.2 &54.6 &64.9 &60.0 &54.4 \\
      \midrule
      \multirow{8}{*}{\rotatebox{90}{CLIP}} & LwF\cite{li2018lwf} & \tr{TPAMI 2017} & 46.8&59.1 &9.4 &25.8 &50.0 &73.9 &79.7 &82.0 &27.1 &26.7 &42.6 &53.5 &52.8 &45.4 \\
      & DKP\textsuperscript{$\dagger$}\cite{xu2024distribution} & \tr{CVPR 2024} &70.2 &77.5 &35.7 &62.1 &80.0 &90.7 &90.9 &91.5 &61.9 &64.0 &67.7 &77.2 &73.9 &66.7 \\
      & LSTKC\textsuperscript{$\dagger$}\cite{xu2024lstkc} & \tr{AAAI 2024} &69.2 &77.6 &34.4 &60.7 &74.9 &88.6 &92.8 &93.7 &60.4 &61.9 &66.3 &76.5 &74.5 &68.2 \\
    & PAEMA\textsuperscript{$\dagger$}\cite{li2024exemplar} & \tr{IJCV 2024} & 71.2 & 79.2 & 32.4 & 58.8 & 78.1 & 90.5 & 92.9 & 93.5 & 57.7 & 58.2 & 66.5 & 76.0 & 75.3 & 68.8
        \\
      
      & DKP++\textsuperscript{$\dagger$}\cite{zhou2025distribution} & \tr{TPAMI 2025} &69.1 &77.5 &35.3 &61.9 &78.6 &89.9 &88.5 &89.2 &58.9 &60.5 &66.1 &75.8 &73.2 &66.6 \\
      & LSTKC++\textsuperscript{$\dagger$}\cite{xu2025long} & \tr{TPAMI 2025} &60.7 &71.0 &34.5 &60.6 &73.3 &87.0 &92.1 &93.0 &70.6 &72.9 &66.2 &76.9 &\silver{76.0} &\silver{69.7} \\
      & DASK\textsuperscript{$\dagger$}\cite{xu2025dask} & \tr{AAAI 2025} &68.7&77.4&39.1
      &65.2&77.7&89.5&91.8
      &92.4&63.7&65.2&\silver{68.2}
      &\silver{77.9}&75.9&68.8\\
      \cmidrule{2-17}
    
      & \textbf{VLADR} & \tr{This Paper} &69.5 &77.4 &40.8 &67.1 &80.6 &91.3 &92.9 &93.6 &67.9 &69.6 &\gold{70.3} &\gold{79.8} &\gold{78.1} &\gold{72.2} \\
      \bottomrule
    \end{tabular}}
\raggedright
 \footnotesize{ Net. represents the Backbone network. $\dag$ indicates integrating the LReID approach with CLIP-ReID~\cite{li2023clip} baseline.}  \\    
    \label{tab:order2}
\end{table*}

\section{Experiments}
\subsection{Datasets and Evaluation Metrics}
\textbf{Datasets}: 
Following previous works~\cite{xu2024lstkc,sun2022patch}, all experiments are conducted on the standard LReID benchmark. Since the DukeMTMC-reID~\cite{ristani2016performance} dataset in the original benchmark has been officially withdrawn, we follow~\cite{xu2025self} and adopt LPW\_s2 as its replacement. The benchmark consists of 12 dataset subsets in total, among which 5 are used for training (Market-1501~\cite{zheng2015scalable}, LPW\_s2~\cite{xu2025self}, CUHK-SYSU ~\cite{xiao2016end}, MSMT17-V2~\cite{wei2018person}, CUHK03~\cite{li2014deepreid}), while the remaining 7 datasets are used exclusively for evaluation (CUHK01~\cite{li2012human}, CUHK02~\cite{li2013locally}, VIPeR~\cite{gray2008viewpoint}, PRID~\cite{hirzer2011person}, i-LIDS~\cite{branch2006imagery}, GRID~\cite{loy2010time},  SenseReID~\cite{zhao2017spindle}).  In accordance with the established LReID protocols \cite{sun2022patch}, two distinct training orders are adopted to simulate different domain gaps~\footnote{(Order-1) Market-1501$\to$CUHK-SYSU$\to$LPW\_s2$\to$MSMT17$\to$ CUHK03}~\footnote{(Order-2) LPW\_s2$\to$MSMT17$\to$Market-1501$\to$CUHK-SYSU$\to$ CUHK03}. The detailed configurations of the datasets are provided in our Supplementary Material. 

Besides, since several closed-source methods have only reported results on the benchmark that contains DukeMTMC-reID, we also provide a comparison with these methods in the Supplementary Material for a comprehensive evaluation.

\textbf{Evaluation Metrics}: Following previous LReID studies~\cite{pu2021lifelong, sun2022patch,xu2024distribution}, we adopt mean Average Precision (mAP) and Rank-1 accuracy (R@1) to evaluate model performance across different domains. In addition, we report the average mAP and R@1 over all training (Seen-Avg) and novel (Unseen-Avg) domains to comprehensively assess the model’s anti-forgetting ability and generalization performance, respectively.

\subsection{Implementation Details}
Our implementation builds upon the framework of~\cite{li2023clip}, utilizing ViT-B/16 as the visual backbone. BLIP-2~\cite{li2023blip} is employed as the image description generator. For MTAD, the training configuration is consistent with~\cite{li2023clip}. For ICAR, the first dataset is trained for 80 epochs, and each subsequent dataset is trained for 60 epochs. The mini-batch size is set to 64, comprising 32 identities with 4 images per identity. All input images are resized to $256 \times 128$. The hyperparameters  $\alpha$ and $\beta$ are both set to 1.0 by default. All experiments are conducted on two NVIDIA RTX 4090 GPUs.

\subsection{Compared Methods}
We compare the proposed VLADR with three categories of LReID models according to their backbone architectures:
A. \textit{ResNet-based models}, including LwF~\cite{li2017learning}, DKP~\cite{xu2024distribution}, LSTKC~\cite{xu2024lstkc}, DKP++~\cite{zhou2025distribution}, LSTKC++~\cite{xu2025long}, and DASK~\cite{xu2025dask};
B. \textit{ViT-based models}, including PAEMA~\cite{li2024exemplar} and DRE~\cite{liu2025diverse};
C. \textit{CLIP-based models}, obtained by integrating the CLIP-ReID baseline with state-of-the-art LReID methods. 
All experimental results are either directly reported from official publications or reproduced using the released implementations. 

The performance of all compared methods is summarized in Tab.~\ref{tab:order1} and Tab.~\ref{tab:order2}, corresponding to training order-1 and order-2, respectively. Results on individual seen domains, average performance across all seen domains (Seen-Avg), and unseen domains (Unseen-Avg) are presented, with the best and second-best results highlighted in \textcolor{red}{\textbf{Red}} and \textcolor{blue}{\textbf{Blue}}, respectively.

\subsection{Seen-Domain Performance Evaluation}
\textbf{Compared to ResNet/ViT-based Methods}: 
As shown in Tab.~\ref{tab:order1} and Tab.~\ref{tab:order2}, VLADR consistently outperforms state-of-the-art ResNet-based and ViT-based approaches across all seen domains. Under Training Order-1, VLADR achieves improvements of \textbf{15.9\%/15.5\%} in Seen-Avg mAP/R@1, and \textbf{10.1\%/9.1\%}. Under Training Order-2, these gains arise because ResNet/ViT-based methods rely on ImageNet-pretrained classification models, whose lack of human-centric priors limits their ability to learn discriminative features. In contrast, our VLM-driven framework effectively leverages the abundant pretrained knowledge embedded in VLMs, which substantially enhances feature extraction during the lifelong learning process.

\textbf{Compare to CLIP-based Methods}: 
As shown in Tab.~\ref{tab:order1} and Tab.~\ref{tab:order2}, integrating existing ResNet/ViT-based approaches into the CLIP-ReID~\cite{li2023clip} baseline generally leads to noticeable performance improvements, with DASK${}^\dagger$ achieving the best results among them due to its historical style data generation mechanism. Nevertheless, VLADR surpasses DASK${}^\dagger$ by \textbf{2.2\%/1.9\%} and \textbf{2.1\%/1.9\%} in Seen-Avg mAP/R@1 across the two training orders. These improvements stem from our attribute disentanglement and reinforcement paradigm, which effectively exploits the human-centric priors within VLMs to enhance human feature parsing and guide the model toward more comprehensive extraction and consolidation of discriminative knowledge across domains.

Furthermore, the learnable text prompts, attribute queries, and local attribute decoder introduced in VLADR are lightweight, adding only marginal computational overhead. In contrast, DASK${}^\dagger$ requires an additional data generation module and significantly enlarges the training set. Therefore, compared with DASK${}^\dagger$, our VLADR is both more \textbf{effective} and more \textbf{efficient}.

\begin{figure}[t]
    \centering
    \includegraphics[width=\linewidth]{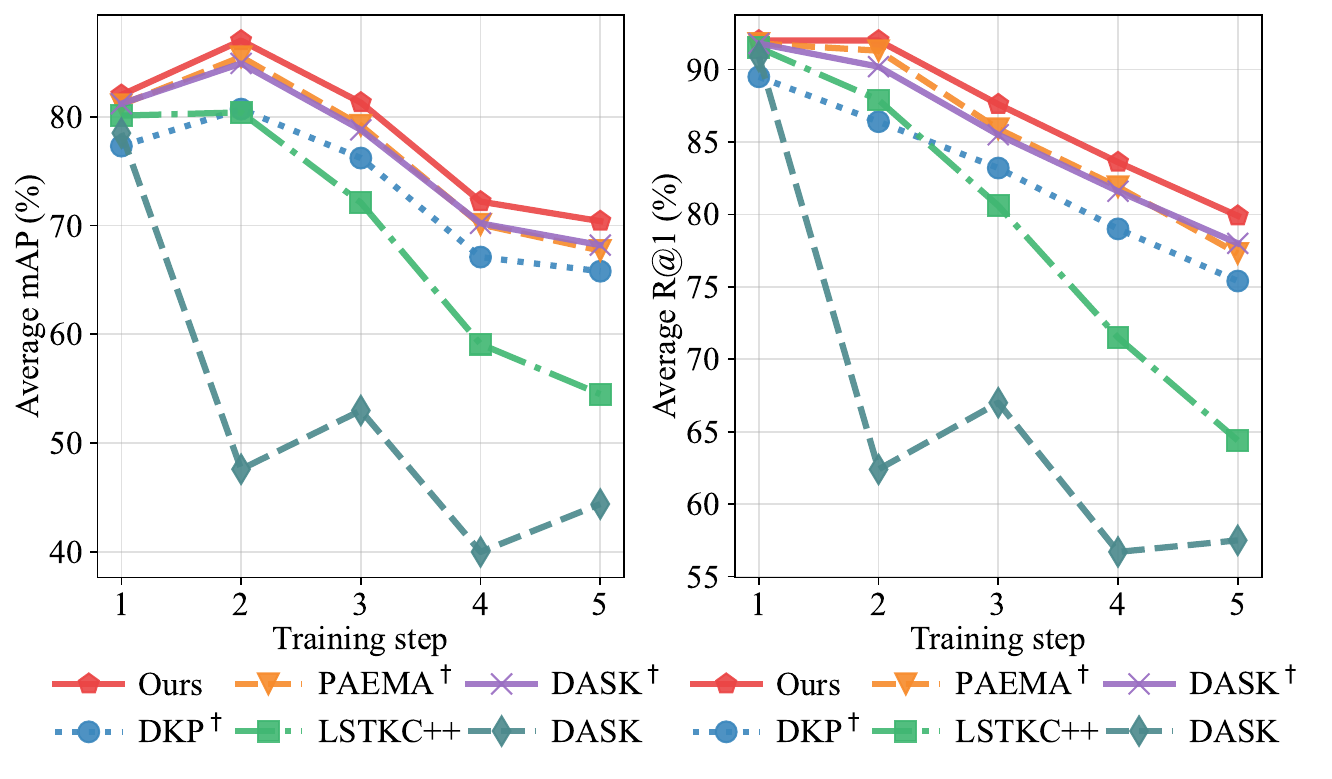}
    \caption{Tendency on seen domain knowledge accumulation.}
    \label{fig:seen_tendency}
\end{figure}

\textbf{Seen-Domain Performance Tendency}:
Fig.~\ref{fig:seen_tendency} illustrates the seen-domain performance trends throughout the sequential training process. At the initial training step, our method exhibits a slight advantage in mAP over all existing approaches. This improvement arises from the attribute disentanglement and visual–text local attribute alignment mechanisms, which enhance the model’s ability to acquire discriminative knowledge. As training proceeds and more domains are introduced, the performance gap between VLADR and existing methods continues to widen. This consistent improvement is primarily attributed to the inter-domain attribute knowledge alignment design, which enables reliable knowledge accumulation and facilitates attribute knowledge reinforcement across domains.

\begin{figure}[t]
    \centering
    \includegraphics[width=\linewidth]{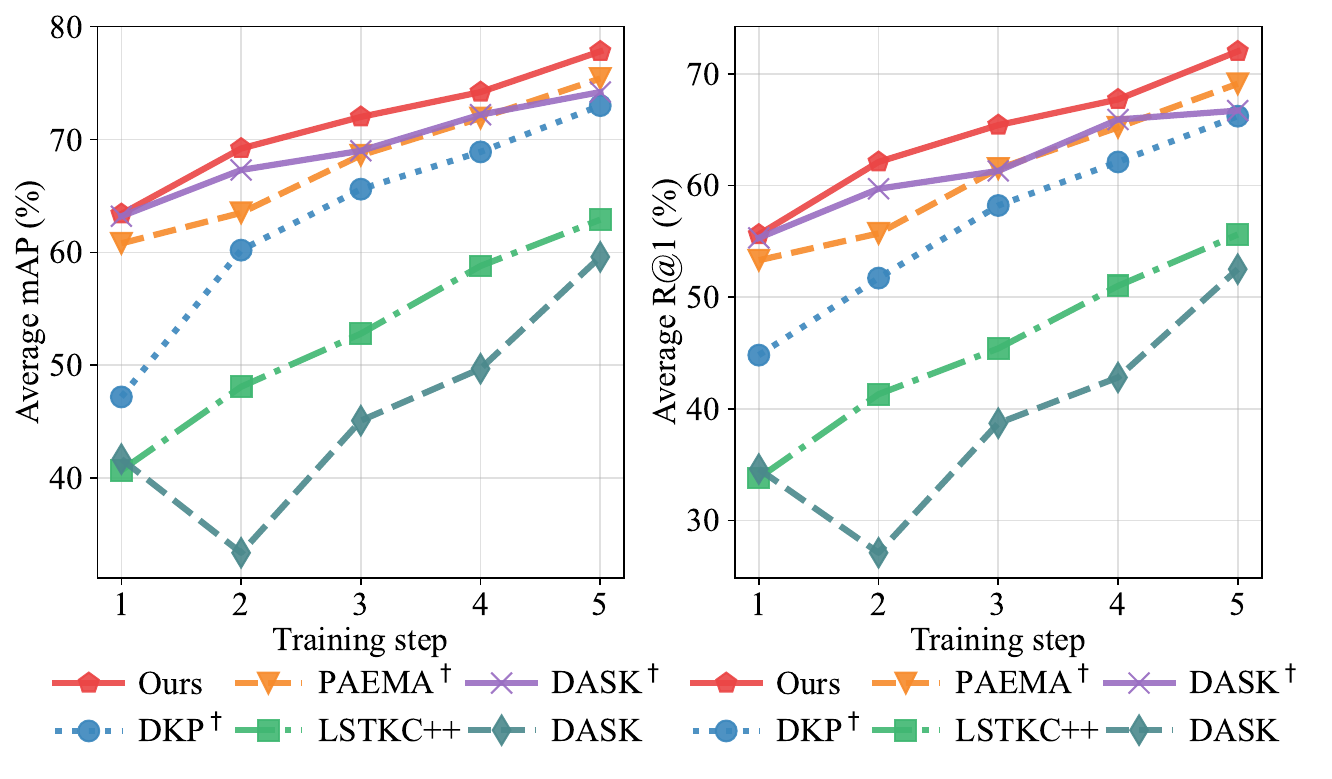}
    \caption{Tendency on unseen domain generalization.}

    \label{fig:unseen_tendency}
\end{figure}

\begin{figure*}[ht]
    \centering
    \includegraphics[width=\linewidth]{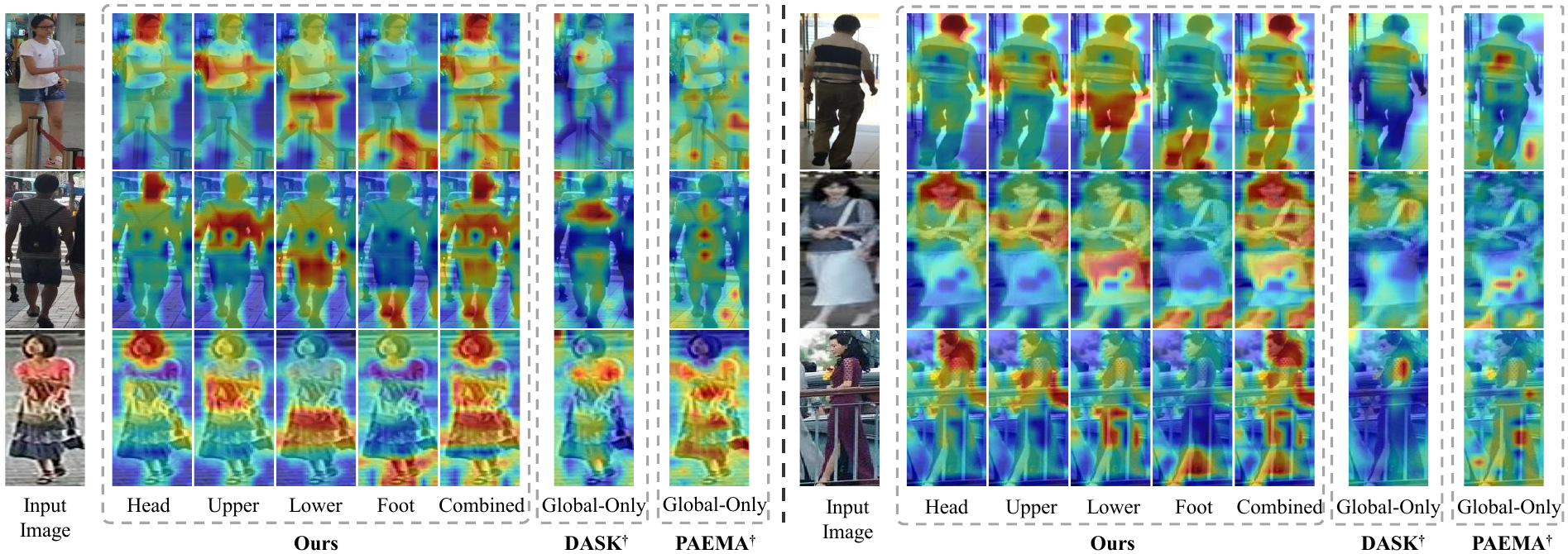}
    \vspace{-10pt}
    \caption{Attention map comparison across different approaches. }
    \vspace{-5pt}
    \label{fig:part_attn_viz}
\end{figure*}

\subsection{Unseen-Domain Generalization Evaluation}
\textbf{Compared to ResNet/ViT/CLIP-based Methods}:
As shown in Tab.~\ref{tab:order1}, VLADR surpasses all compared methods by at least \textbf{14.9\%/16.4\%} in Unseen-Avg mAP/R@1 when compared to ResNet-based models, and by \textbf{8.0\%/8.8\%} relative to ViT-based approaches. For CLIP-based models, the minimum improvement is \textbf{2.2\%/2.2\%}. Likewise, under Training Order-2 (Tab.~\ref{tab:order2}), VLADR consistently outperforms all ResNet/ViT/CLIP-based methods, achieving margins of at least \textbf{2.1\%/2.5\%} in Unseen-Avg mAP/R@1. These substantial improvements stem from the proposed attribute disentanglement and reinforcement mechanisms, which guide the model to learn universally discriminative and domain-invariant features that generalize effectively to unseen environments.

\textbf{Unseen-Domain Performance Tendency}:
As presented in Fig.~\ref{fig:unseen_tendency}, our VLADR exhibits comparable generalization ability to DASK${}^\dagger$ at the initial training step, and consistently outperforms all existing methods as training progresses. This trend demonstrates that the proposed vision–language attribute disentanglement and reinforcement paradigm enables the continual accumulation of universal, transferable knowledge throughout the learning process.

\textbf{Attention Map visualization}: 
To further demonstrate the superiority of the proposed fine-grained attribute modeling mechanism, we visualize the attention maps generated by each local attribute query, as well as their combined attention map. For comparison, global attention maps from existing approaches are also included. The results in Fig.~\ref{fig:part_attn_viz} show that each local attribute query accurately captures part-level semantics, and their combination yields a comprehensive representation of human information. In contrast, existing methods highlight only partial human regions or undesirably attend to background areas, reflecting their limited capacity for precise semantic extraction.

\begin{table}[t]
    \centering
    \caption{Ablation on the proposed model components.}
    \vspace{-5pt}
    \setlength{\tabcolsep}{1.2mm}{\begin{tabular}{ccccccc}
    \toprule
    \multirow{2}{*}{Baseline} & \multirow{2}{*}{$\mathcal{L}_{MAlign}$} & \multirow{2}{*}{$\mathcal{L}_{DAlign}$} & \multicolumn{2}{c}{Seen-Avg} & \multicolumn{2}{c}{Unseen-Avg} \\
     & & & mAP & R@1 & mAP & R@1 \\
     \midrule
     $\checkmark$ & && 57.1 & 66.6 & 66.4 & 58.8 \\
     $\checkmark$& $\checkmark$ && 66.3 & 75.8 & 74.6 & 68.0 \\
     $\checkmark$ & $\checkmark$ & $\checkmark$ & \textbf{70.4} & \textbf{79.9} & \textbf{77.8} & \textbf{72.0} \\
     \bottomrule
    \end{tabular}}    
    \vspace{-5pt}
    \label{tab:ablation_component}
\end{table}

\subsection{Ablation Studies}
We perform ablation studies on the components proposed under Training Order-1 as follows:

\textbf{Ablations on model components}: 
As shown in Tab.~\ref{tab:ablation_component}, the baseline is the CLIP-ReID~\cite{li2023clip} framework trained with $\mathcal{L}_{ReID}+\mathcal{L}_{global}$. When incorporating $\mathcal{L}_{MAlign}$, both the Multi-grain Text Attribute Disentanglement module and $\mathcal{L}_{MAlign}$ are applied to facilitate visual local attribute mining. This introduces substantial improvements of \textbf{9.2\%/9.2\%} in Seen-Avg mAP/R@1 and \textbf{8.2\%/9.2\%} in Unseen-Avg mAP/R@1. These results validate that the proposed human attribute disentanglement design significantly enhances discriminative feature learning.
Furthermore, when $\mathcal{L}_{DAlign}$ is additionally employed, VLADR achieves \textbf{13.3\%/13.3\%} gains in Seen-Avg mAP/R@1 and \textbf{11.4\%/13.2\%} gains in Unseen-Avg mAP/R@1 over the baseline. These improvements demonstrate that the proposed continual attribute knowledge reinforcement mechanism effectively improves both domain adaptability and cross-domain generalization.

\begin{figure}[t]
    \centering
    \includegraphics[width=\linewidth,trim=0cm 0.2cm 0cm 0.2cm,clip]{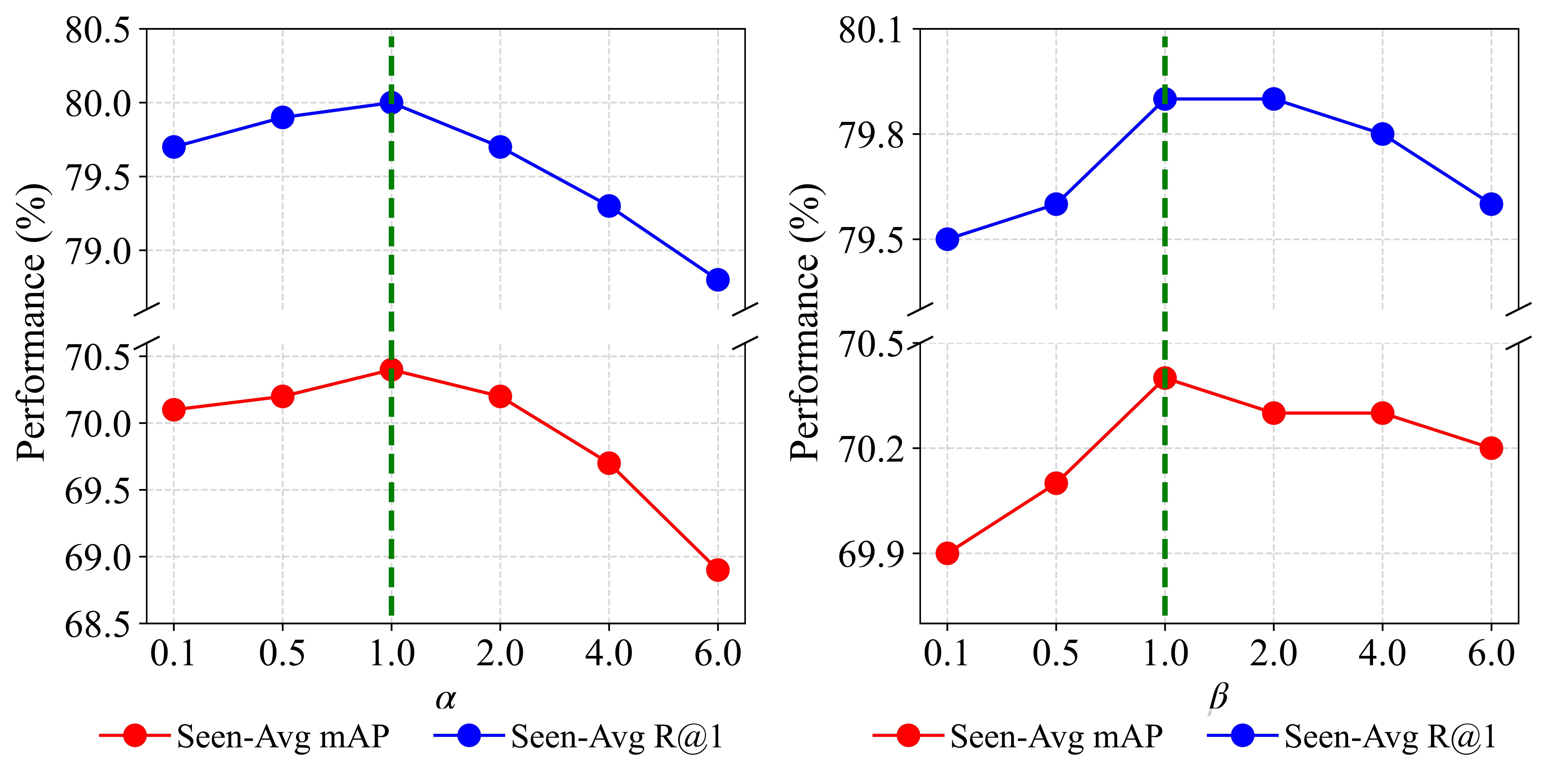}
    \caption{Ablation on hyperparameters.}
    \vspace{-10pt}
    \label{fig:hyper_ablation}
\end{figure}

\textbf{Ablation on hyperparamerters}:
We further investigate the sensitivity of VLADR to the hyperparameters $\alpha$ and $\beta$, as shown in Fig.~\ref{fig:hyper_ablation}. The results indicate that the model maintains stable performance across a wide range of values, demonstrating strong robustness to hyperparameter fluctuations. In practice, we set both $\alpha$ and $\beta$ to 1.0 by default based on empirical observations.



\section{Conclusion}
This paper presents VLADR, a novel VLM-driven LReID framework. The central idea is to explicitly model universally shared human attributes to improve both intra-domain knowledge learning and inter-domain knowledge transfer, thereby forming a continual attribute knowledge reinforcement paradigm.Specifically, VLADR introduces a Multi-grain Text Attribute Disentanglement mechanism to extract comprehensive global and diverse local textual attributes. Additionally, an Inter-domain Cross-modal Attribute Reinforcement scheme jointly performs cross-modal and inter-domain attribute alignment, enabling effective visual attribute mining and robust knowledge consolidation. This work demonstrates the strong potential of attribute-centric modeling in LReID and provides an efficient solution for fine-grained attribute knowledge discovery without imposing heavy annotation overhead.


 \textbf{Acknowledgements}:
This work was a research achievement of National Engineering Research Center of New Electronic Publishing Technologies, and was supported by the National Key R\&D Program of China (2024YFA1410000), the National Natural Science Foundation of China (62376011, 62406313), and the Postdoctoral Fellowship Program of China Postdoctoral Science Foundation, Grant No. YJB20250283.

{
    \small
    \bibliographystyle{ieeenat_fullname}
    \bibliography{main}

@String(PAMI = {IEEE Trans. Pattern Anal. Mach. Intell.})

@String(IJCV = {Int. J. Comput. Vis.})

@String(CVPR= {IEEE Conf. Comput. Vis. Pattern Recog.})

@String(ICCV= {Int. Conf. Comput. Vis.})

@String(ECCV= {Eur. Conf. Comput. Vis.})

@String(ACMMM= {ACM Int. Conf. Multimedia})

@String(ACCV  = {ACCV})

@String(AAAI = {AAAI})

@String(CVPRW= {IEEE Conf. Comput. Vis. Pattern Recog. Worksh.})

@String(CSVT = {IEEE Trans. Circuit Syst. Video Technol.})

@String(PAMI  = {IEEE TPAMI})

@String(IJCV  = {IJCV})

@String(CVPR  = {CVPR})

@String(ICCV  = {ICCV})

@String(ECCV  = {ECCV})

@String(NeurIPS  = {NeurIPS})

@String(ICML  = {ICML})

@String(ACMMM = {ACM MM})

@String(CVPRW= {CVPRW})

@String(CSVT = {IEEE TCSVT})

@inproceedings{wu2021generalising,
  title={Generalising without Forgetting for Lifelong Person Re-identification},
  author={Wu, Guile and Gong, Shaogang},
  booktitle=AAAI,
  pages={2889--2897},
  year={2021}
}

@inproceedings{pu2021lifelong,
  title={Lifelong Person Re-Identification via Adaptive Knowledge Accumulation},
  author={Pu, Nan and Chen, Wei and Liu, Yu and Bakker, Erwin M and Lew, Michael S},
  booktitle=CVPR,
  pages={7897--7906},
  year={2021},
  organization={IEEE}
}

@inproceedings{zheng2015scalable,
  title={Scalable person re-identification: A benchmark},
  author={Zheng, Liang and Shen, Liyue and Tian, Lu and Wang, Shengjin and Wang, Jingdong and Tian, Qi},
  booktitle=ICCV,
  pages={1116--1124},
  year={2015},
  organization={IEEE}
}

@inproceedings{ristani2016performance,
  title={Performance measures and a data set for multi-target, multi-camera tracking},
  author={Ristani, Ergys and Solera, Francesco and Zou, Roger and Cucchiara, Rita and Tomasi, Carlo},
  booktitle=ECCV,
  pages={17--35},
  year={2016},
  organization={Springer}
}

@inproceedings{luo2019bag,
  title={Bag of tricks and a strong baseline for deep person re-identification},
  author={Luo, Hao and Gu, Youzhi and Liao, Xingyu and Lai, Shenqi and Jiang, Wei},
  booktitle=CVPRW,
  pages={0--0},
  year={2019}
}

@inproceedings{zhuang2020rethink,
    title={Rethinking the Distribution Gap of Person Re-identification with Camera-based Batch Normalization},
    author={Zijie Zhuang and Longhui Wei and Lingxi Xie and Tianyu Zhang and Hengheng Zhang and Haozhe Wu and Haizhou Ai and Qi Tian},
    year={2020},
    organization={Springer},
    booktitle=ECCV,
    pages={140--157}
}

@INPROCEEDINGS{he2021transreid,
  author={He, Shuting and Luo, Hao and Wang, Pichao and Wang, Fan and Li, Hao and Jiang, Wei},
  booktitle=ICCV, 
  title={TransReID: Transformer-based Object Re-Identification}, 
  year={2021},
  pages={14993-15002},
    organization={IEEE}
}

@inproceedings{li2018HAN,
title = {Harmonious Attention Network for Person Re-Identification},
author = {Li, Wei and Zhu, Xiatian and Gong, Shaogang},
booktitle = CVPR,
pages = {2285-2294},
year = {2018},
organization = {IEEE}
}

@misc{xiao2016end,
  title={End-to-end deep learning for person search},
  author={Xiao, Tong and Li, Shuang and Wang, Bochao and Lin, Liang and Wang, Xiaogang},
eprint={1604.01850},
  archivePrefix={arXiv},
  volume={2},
  number={2},
  pages={4},
  year={2016}
}

@inproceedings{sun2022patch,
  title={Patch-based Knowledge Distillation for Lifelong Person Re-Identification},
  author={Sun, Zhicheng and Mu, Yadong},
  booktitle=ACMMM,
  pages={696--707},
  year={2022}
}

@inproceedings{huang2022lifelong,
  title={Lifelong Unsupervised Domain Adaptive Person Re-identification with Coordinated Anti-forgetting and Adaptation},
  author={Huang, Zhipeng and Zhang, Zhizheng and Lan, Cuiling and Zeng, Wenjun and Chu, Peng and You, Quanzeng and Wang, Jiang and Liu, Zicheng and Zha, Zheng-jun},
  booktitle=CVPR,
  pages={14288--14297},
  year={2022},
  organization={IEEE}
}

@inproceedings{yu2023lifelong,
  title={Lifelong Person Re-Identification via Knowledge Refreshing and Consolidation},
  author={Yu, Chunlin and Shi, Ye and Liu, Zimo and Gao, Shenghua and Wang, Jingya},
  booktitle=AAAI,
  pages={3295--3303},
  year={2023}
}

@inproceedings{ge2022lifelong,
  title={Lifelong Person Re-identification by Pseudo Task Knowledge Preservation},
  author={Ge, Wenhang and Du, Junlong and Wu, Ancong and Xian, Yuqiao and Yan, Ke and Huang, Feiyue and Zheng, Wei-Shi},
  booktitle=AAAI,
  pages={688--696},
  year={2022}
}

@article{li2017learning,
  title={Learning without Forgetting},
  author={Li, Zhizhong and Hoiem, Derek},
  journal=PAMI,
  volume={40},
  number={12},
  pages={2935--2947},
  year={2017},
  publisher={IEEE}
}

@ARTICLE{chen2018camera,
  author={Chen, Ying-Cong and Zhu, Xiatian and Zheng, Wei-Shi and Lai, Jian-Huang},
  journal=PAMI, 
  title={Person Re-Identification by Camera Correlation Aware Feature Augmentation}, 
  year={2018},
  volume={40},
  number={2},
  pages={392-408}
}

@ARTICLE{li2018lwf,
  author={Li, Zhizhong and Hoiem, Derek},
  journal=PAMI, 
  title={Learning without Forgetting}, 
  year={2018},
  volume={40},
  number={12},
  pages={2935-2947},
}

@INPROCEEDINGS{Shemelkov2017IL,
  title={Incremental Learning of Object Detectors without Catastrophic Forgetting}, 
  author={Shmelkov, Konstantin and Schmid, Cordelia and Alahari, Karteek},
  booktitle=ICCV, 
  year={2017},
  pages={3420-3429}
}

@article{liu2025tp,
  title={TP-LReID: Lifelong Person Re-identification using Text Prompts},
  author={Liu, Zhaoshuo and Guo, Zhiwei and Feng, Chaolu and Li, Wei and Yu, Kun and Hu, Jun and Yang, Jinzhu},
  journal={Pattern Recognition},
  pages={112326},
  year={2025},
  publisher={Elsevier}
}

@article{wang2024large,
  title={When large vision-language models meet person re-identification},
  author={Wang, Qizao and Li, Bin and Xue, Xiangyang},
  journal={arXiv preprint arXiv:2411.18111},
  year={2024}
}

@article{yang2024mllmreid,
  title={MLLMReID: multimodal large language model-based person re-identification},
  author={Yang, Shan and Zhang, Yongfei},
  journal={arXiv preprint arXiv:2401.13201},
  year={2024}
}

@inproceedings{li2012human,
  title={Human Reidentification with Transferred Metric Learning},
  author={Li, Wei and Zhao, Rui and Wang, Xiaogang},
  booktitle=ACCV,
  pages={31--44},
  year={2012},
  organization={Springer}
}

@inproceedings{li2014deepreid,
  title={DeepReID: Deep Filter Pairing Neural Network for Person Re-identification},
  author={Li, Wei and Zhao, Rui and Xiao, Tong and Wang, Xiaogang},
  booktitle=CVPR,
  pages={152--159},
  year={2014},
  organization={IEEE}
}

@article{loy2010time,
  title={Time-delayed Correlation Analysis for Multi-camera Activity Understanding},
  author={Loy, Chen Change and Xiang, Tao and Gong, Shaogang},
  journal=IJCV,
  volume={90},
  number={1},
  pages={106--129},
  year={2010},
  publisher={Springer}
}

@inproceedings{zhao2017spindle,
  title={Spindle Net: Person Re-identification with Human Body Region Guided Feature Decomposition and Fusion},
  author={Zhao, Haiyu and Tian, Maoqing and Sun, Shuyang and Shao, Jing and Yan, Junjie and Yi, Shuai and Wang, Xiaogang and Tang, Xiaoou},
  booktitle=CVPR,
  pages={907--915},
  year={2017},
  organization={IEEE}
}

@article{liu2025diverse,
  title={Diverse representations embedding for lifelong person re-identification},
  author={Liu, Shiben and Fan, Huijie and Wang, Qiang and Chen, Xiai and Han, Zhi and Tang, Yandong},
  journal={IEEE Transactions on Neural Networks and Learning Systems},
  year={2025},
  publisher={IEEE}
}

@inproceedings{li2013locally,
  title={Locally aligned feature transforms across views},
  author={Li, Wei and Wang, Xiaogang},
  booktitle=CVPR,
  pages={3594--3601},
  year={2013},
  organization={IEEE}
}

@inproceedings{gray2008viewpoint,
  title={Viewpoint invariant pedestrian recognition with an ensemble of localized features},
  author={Gray, Douglas and Tao, Hai},
  booktitle=ECCV,
  pages={262--275},
  year={2008},
  organization={Springer}
}

@inproceedings{hirzer2011person,
  title={Person re-identification by descriptive and discriminative classification},
  author={Hirzer, Martin and Beleznai, Csaba and Roth, Peter M and Bischof, Horst},
  booktitle={Image Analysis},
  pages={91--102},
  year={2011},
  organization={Springer}
}

@inproceedings{branch2006imagery,
  title={Imagery library for intelligent detection systems (i-lids)},
  author={Branch, Home Office Scientific Development},
  booktitle={2006 IET conference on crime and security},
  pages={445--448},
  year={2006},
  organization={IET}
}

@inproceedings{xu2024distribution,
  title={Distribution-aware Knowledge Prototyping for Non-exemplar Lifelong Person Re-identification},
  author={Xu, Kunlun and Zou, Xu and Peng, Yuxin and Zhou, Jiahuan},
  booktitle=CVPR,
  pages={16604--16613},
  year={2024},
  organization={IEEE}
}

@inproceedings{xu2024lstkc,
  title={LSTKC: Long Short-Term Knowledge Consolidation for Lifelong Person Re-identification},
  author={Xu, Kunlun and Zou, Xu and Zhou, Jiahuan},
  booktitle=AAAI,
  volume={38},
  number={14},
  pages={16202--16210},
  year={2024}
}

@inproceedings{wang2022dualprompt,
  title={Dualprompt: Complementary prompting for rehearsal-free continual learning},
  author={Wang, Zifeng and Zhang, Zizhao and Ebrahimi, Sayna and Sun, Ruoxi and Zhang, Han and Lee, Chen-Yu and Ren, Xiaoqi and Su, Guolong and Perot, Vincent and Dy, Jennifer and others},
  booktitle=ECCV,
  pages={631--648},
  year={2022},
  organization={Springer}
}

@inproceedings{xu2024mitigate,
  title={Mitigate Catastrophic Remembering via Continual Knowledge Purification for Noisy Lifelong Person Re-Identification},
  author={Xu, Kunlun and Zhang, Haozhuo and Li, Yu and Peng, Yuxin and Zhou, Jiahuan},
  booktitle=ACMMM,
pages={5790--5799},
  year={2024}
}

@inproceedings{shi2023dual,
  title={Dual pseudo-labels interactive self-training for semi-supervised visible-infrared person re-identification},
  author={Shi, Jiangming and Zhang, Yachao and Yin, Xiangbo and Xie, Yuan and Zhang, Zhizhong and Fan, Jianping and Shi, Zhongchao and Qu, Yanyun},
  booktitle=ICCV,
  pages={11218--11228},
  year={2023},
  publisher={IEEE}
}

@inproceedings{cui2024learning,
  title={Learning continual compatible representation for re-indexing free lifelong person re-identification},
  author={Cui, Zhenyu and Zhou, Jiahuan and Wang, Xun and Zhu, Manyu and Peng, Yuxin},
  booktitle=CVPR,
  pages={16614--16623},
  year={2024}
}

@inproceedings{xu2025dask,
  title={Dask: Distribution rehearsing via adaptive style kernel learning for exemplar-free lifelong person re-identification},
  author={Xu, Kunlun and Jiang, Chenghao and Xiong, Peixi and Peng, Yuxin and Zhou, Jiahuan},
  booktitle=AAAI,
  volume={39},
  number={9},
  pages={8915--8923},
  year={2025}
}

@article{xu2025long,
  title={Long Short-Term Knowledge Decomposition and Consolidation for Lifelong Person Re-Identification},
  author={Xu, Kunlun and Liu, Zichen and Zou, Xu and Peng, Yuxin and Zhou, Jiahuan},
  journal=PAMI,
  year={2025},
  publisher={IEEE}
}

@inproceedings{cui2025dkc,
  title={DKC: Differentiated Knowledge Consolidation for Cloth-Hybrid Lifelong Person Re-identification},
  author={Cui, Zhenyu and Zhou, Jiahuan and Peng, Yuxin},
  booktitle=CVPR,
  pages={3573--3582},
  year={2025}
}

@article{li2024exemplar,
  title={Exemplar-free lifelong person re-identification via prompt-guided adaptive knowledge consolidation},
  author={Li, Qiwei and Xu, Kunlun and Peng, Yuxin and Zhou, Jiahuan},
  journal=IJCV,
  volume={132},
  number={11},
  pages={4850--4865},
  year={2024},
  publisher={Springer}
}

@article{zhou2025distribution,
  title={Distribution-Aware Knowledge Aligning and Prototyping for Non-Exemplar Lifelong Person Re-Identification},
  author={Zhou, Jiahuan and Xu, Kunlun and Zhuo, Fan and Zou, Xu and Peng, Yuxin},
  journal=PAMI,
  year={2025},
  publisher={IEEE}
}

@inproceedings{xu2025self,
  title={Self-Reinforcing Prototype Evolution with Dual-Knowledge Cooperation for Semi-Supervised Lifelong Person Re-Identification},
  author={Xu, Kunlun and Zhuo, Fan and Li, Jiangmeng and Zou, Xu and Jiahuan Zhou},
  booktitle=ICCV,
  year={2025}
}

@article{cui2024dma,
  title={DMA: Dual modality-aware alignment for visible-infrared person re-identification},
  author={Cui, Zhenyu and Zhou, Jiahuan and Peng, Yuxin},
  journal=TIFS,
  volume={19},
  pages={2696--2708},
  year={2024},
  publisher={IEEE}
}

@article{cui2023dcr,
  title={Dcr-reid: Deep component reconstruction for cloth-changing person re-identification},
  author={Cui, Zhenyu and Zhou, Jiahuan and Peng, Yuxin and Zhang, Shiliang and Wang, Yaowei},
  journal=CSVT,
  volume={33},
  number={8},
  pages={4415--4428},
  year={2023},
  publisher={IEEE}
}

@inproceedings{zhang2025scap,
  title={Scap: Transductive test-time adaptation via supportive clique-based attribute prompting},
  author={Zhang, Chenyu and Xu, Kunlun and Liu, Zichen and Peng, Yuxin and Zhou, Jiahuan},
  booktitle=CVPR,
  pages={30032--30041},
  year={2025}
}

@inproceedings{wei2018person,
  title={Person Transfer GAN to Bridge Domain Gap for Person Re-identification},
  author={Wei, Longhui and Zhang, Shiliang and Gao, Wen and Tian, Qi},
  booktitle=CVPR,
  pages={79--88},
  year={2018},
  organization={IEEE}
}

@inproceedings{xu2025componential,
  title={Componential Prompt-Knowledge Alignment for Domain Incremental Learning},
  author={Xu, Kunlun and Zou, Xu and Hua, Gang and Zhou, Jiahuan},
  booktitle=ICML, 
  year={2025}
}

@inproceedings{gong2022person,
  title={Person re-identification method based on color attack and joint defence},
  author={Gong, Yunpeng and Huang, Liqing and Chen, Lifei},
  booktitle=CVPR,
  pages={4313--4322},
  year={2022}
}

@article{gong2024cross,
  title={Cross-modality perturbation synergy attack for person re-identification},
  author={Gong, Yunpeng and Zhong, Zhun and Qu, Yansong and Luo, Zhiming and Ji, Rongrong and Jiang, Min},
  journal=NeurIPS,
  volume={37},
  pages={23352--23377},
  year={2024}
}

@article{hu2024empowering,
  title={Empowering visible-infrared person re-identification with large foundation models},
  author={Hu, Zhangyi and Yang, Bin and Ye, Mang},
  journal=NeurIPS,
  volume={37},
  pages={117363--117387},
  year={2024}
}

@inproceedings{li2023clip,
  title={Clip-reid: exploiting vision-language model for image re-identification without concrete text labels},
  author={Li, Siyuan and Sun, Li and Li, Qingli},
  booktitle=AAAI,
  volume={37},
  number={1},
  pages={1405--1413},
  year={2023}
}

@inproceedings{yang2024pedestrian,
  title={A pedestrian is worth one prompt: Towards language guidance person re-identification},
  author={Yang, Zexian and Wu, Dayan and Wu, Chenming and Lin, Zheng and Gu, Jingzi and Wang, Weiping},
  booktitle=CVPR,
  pages={17343--17353},
  year={2024}
}

@article{mao2023clip4hoi,
  title={Clip4hoi: Towards adapting clip for practical zero-shot hoi detection},
  author={Mao, Yunyao and Deng, Jiajun and Zhou, Wengang and Li, Li and Fang, Yao and Li, Houqiang},
  journal=NeurIPS,
  volume={36},
  pages={45895--45906},
  year={2023}
}

@inproceedings{chen2023unveiling,
  title={Unveiling the power of clip in unsupervised visible-infrared person re-identification},
  author={Chen, Zhong and Zhang, Zhizhong and Tan, Xin and Qu, Yanyun and Xie, Yuan},
  booktitle={ACM MM},
  pages={3667--3675},
  year={2023}
}

@inproceedings{li2023blip,
  title={Blip-2: Bootstrapping language-image pre-training with frozen image encoders and large language models},
  author={Li, Junnan and Li, Dongxu and Savarese, Silvio and Hoi, Steven},
  booktitle={ICML},
  pages={19730--19742},
  year={2023},
  organization={PMLR}
}
}


\end{document}